\documentclass[twoside,11pt]{article}

\usepackage{template, natbib, rawfonts}
\usepackage{graphicx}
\usepackage{amsmath, amsfonts, amssymb, float, fancyhdr}
\usepackage{xcolor}
\usepackage{algorithm}
\usepackage{mathptmx}
\usepackage{algorithm}
\usepackage{algorithmic}
\usepackage{multirow}
\usepackage{amssymb}
\usepackage{amsthm,thmtools}
\usepackage{breqn}
\usepackage{makecell}

\theoremstyle{asmp}
\newtheorem{asmp}{Assumption}[section]
\theoremstyle{lemma}
\newtheorem{lemma}{Lemma}[section]
\DeclareUnicodeCharacter{2212}{\textendash}
\usepackage[colorlinks=false,allbordercolors={1 1 1}]{hyperref}

\begin{document}
\title{Grad Queue : A probabilistic framework to reinforce sparse gradients}
\author{\name Irfan Mohammad Al Hasib\\ \email irfanhasib.me@gmail.com}


\maketitle

\begin{abstract}
   Informative gradients are often lost in large batch updates. We propose a robust mechanism to reinforce the sparse components within a random batch of data points. A finite queue of online gradients is used to determine their expected instantaneous statistics. We propose a function to measure the scarcity of incoming gradients using these statistics and establish the theoretical ground of this mechanism. To minimize conflicting components within large mini-batches, samples are grouped with aligned objectives by clustering based on inherent feature space. Sparsity is measured for each centroid and weighted accordingly. A strong intuitive criterion to squeeze out redundant information from each cluster is the backbone of the system. It makes rare information indifferent to aggressive momentum also exhibits superior performance with larger mini-batch horizon. The effective length of the queue kept variable to follow the local loss pattern. The contribution of our method is to restore intra-mini-batch diversity at the same time widening the optimal batch boundary. Both of these collectively drive it deeper towards the minima. Our method has shown superior performance for CIFAR10, MNIST, and Reuters News category dataset compared to mini-batch gradient descent.

\end{abstract}

\section{Introduction}
\label{Introduction}
Pursuing the global minima is a fundamental problem in gradient based learning. Stochastic and batch gradient descent are the two opposite extremes in this regime. Stochastic gradient descent suffers from the inherent sample variance \cite{johnson2013accelerating,fang2018spider,wang2018spiderboost,wang2019spiderboost}. In contrast full batch gradient descent has different problems i.e - (i) computation cost per update for large datasets (ii) poor performance due to generalization gap. Computational cost can be counteracted by distributed learning \cite{goyal2017accurate}. The reason of generalization gap is not well defined yet. Apart from over-fitting, \cite{keskar2016large} addresses lack of exploratory property and attraction to saddle points as reasons. \cite{yin2018gradient} addresses lack of diversity in consecutive updates as a potential reason. Informative gradients can disappear in large mini-batch causing reduced diversity. In this work we aim to resolve this by boosting the rare updates. 
Availability of large GPU (graphical processing unit) memory and distributed learning leads to unprecedented growth in batch size with reasonable time constraint. Therefore, pushing the upper bound of mini-batch is one of the prime focus in deep learning research \cite{goyal2017accurate,yin2018gradient,de2016big}.\\
 Monotonous information keeps repeating thus often learnt quickly, leaving rare information the prime key to dive lower in the loss curve. The success of the adaptive optimizers relies on this fact \cite{duchi2011adaptive} \cite{hinton2012neural} \cite{kingma2014adam}. The common goal of all these algorithms is to extract the most informative updates from a set of gradients. To aid this objective, we detect the sparse updates by quantifying its distance from magnitude spectrum in the recent past. We append an amplification plugin on top of any arbitrary gradient acquisition pipeline. It amplify the sparse and minimize the monotonous values, resulting a good contrast. The margin of this reinforced diversity is controlled inherently by their variance. The larger the batch size gets the more invisible the sparse signal becomes, our method come into rescue. It demonstrates the potential to increase accuracy at optimal batch size and specially beyond it. We weight the new updates by its scarcity based on past trend from the queue. For beyond optimal batch size, we group them using k-means clustering (based on a latent feature space). Then extract and emphasize the cluster centers with highest informativeness before totaling them at the end. Informativeness is determined by the scarcity of the occurrence. This intra-mini-batch grouping and enforcing the sparse components per group, ensures minimal loss of informative gradients. Lower risk of loosing diversity allows larger batch size. In case of a single cluster this method will still boost the rarely occurring signals to aid better utilization of the unique updates. The length of the queue is updated based on the trend in change of loss value for the past updates. By making the queue length short and flexible we focus on the subset of the past gradients which can help the current update most. Incorporating a probabilistic approach makes it more suitable for the stochastic nature of the process. We also classify monotonous, sparse and noisy updates and their effect on mini-batch update for better understanding our approach.

\section{Literature review}
\label{network}
Extracting the most out of a particular set of data points is a fundamental desideratum in machine learning. For batch update, works include optimizing the size of a mini-batch, training with larger batch size. In contrast, for online gradient descent variance reduction techniques are applied for fast convergence. Some works aim to design efficient data selection strategy for each update for getting more useful gradients. While others try to manipulate the gradient itself by comparing its distribution from a auxiliary sources. In this work we are comparing the gradient with its own past distribution.\\
\textbf{Size of Mini-batch :} Optimization difficulty caused by large mini-batch is the prime obstacle in speeding up training with large datasets. The speed up saturation with distributed learning drawing more attention to this  phenomenon \cite{keskar2016large}\cite{goyal2017accurate} proposes several strategies to overcome it. \cite{yin2018gradient} proposed a lower optimal bound for batch size to maintain the diversity of updates . \cite{friedlander2012hybrid}, \cite{de2016big} attempts to resolve it by adaptively growing batch size with time. \cite{goyal2017accurate} proposes several workaround to push the upper limit of batch size. Yet today it is left to be tuned real time by the practitioners by and large.\\
\textbf{Variance Reduction:} Instead of going for larger batch size,  many works propose to reduce the variance of stochistic update.  \cite{johnson2013accelerating}\cite{fang2018spider}\cite{wang2018spiderboost},\cite{wang2019spiderboost} proposed to reduce variance by keeping a snapshot of older weights in the expense of a complete pass through the entire dataset. \cite{elibol2020variance} proposes special operator to minimize this computational cost.\\
\textbf{Sample selection strategy:} These approaches attempt to improve the quality of the mini batch by creating a dataset with informative samples regardless of the size of mini-batch.\cite{agarwal2022estimating},\cite{zhdanov2019diverse},\cite{csiba2018importance}. Some works focus on serving the samples in phases based on their inherent complexity \cite{bengio2009curriculum} \cite{alain2015variance} \cite{jiang2018mentornet} \cite{fan2017learning} \cite{kumar2010self} \cite{loshchilov2015online}.\\ 
\textbf{Gradient manipulation :} Auxiliary task learning aims to preserve the important components of the gradients by comparing their distribution with a separate source of gradients from an auxiliary task \cite{al2023boosting} \cite{du2018adapting}. In that sense our work compares the gradients with its own past distribution within a close horizon. Clustering the gradients can minimize the variance within the clusters thus reduces destructive interference while averaging. \cite{faghri2020study} proposed a learnable framework for the cluster centers in expense of additional computation. In this work we cluster the samples thus the gradients, based on the activations from a dense layer. The use of dense outputs from pre-trained model for image matching or retrieval \cite{efe2021dfm} \cite{arandjelovic2016netvlad} inspire us to use dense features for grouping the gradients.

\section{Methodology}
\begin{figure}[htbp]
\vskip 0in
\begin{center}
\centerline{\includegraphics[width=8cm]{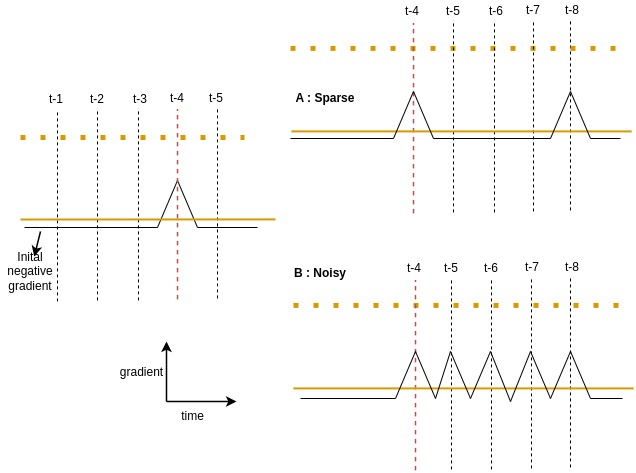}}
\caption{Figure 2: }
\label{fig1}
\end{center}
\vskip -0in
\end{figure}
\subsection{Attention to rare signal}
From a information science perspective, a useful signal is something that adds some unique value to the process. In that sense signal to noise ratio can be controlled little counter-intuitively by paying attention to the rare spikes, similar to a low pass filter. This hypothesis holds for online gradient acquisition in stochastic method. There is no guaranteed way to distinguish between a useful sparse signal and series of noisy spikes from the first appearance. Its nature can be established based on later occurrences. The more we observe the future values the more it will be clearer. fig-[\ref{fig1}] One unavoidable issue in stochastic optimization is that it cannot know the upcoming gradients in advance. At best, the past trend can be stored and analyzed to determine the consistency of the current signal. We acknowledge the exponential averaging methods \cite{kingma2014adam, duchi2011adaptive}] can preserve  the sparsity among gradients to some extant by making a balance between first and second moment. Here we rather aim to design a standalone distance function to compute a numeric measure.

We measure the instantaneous expected value and standard deviation for each gradient from the queue. The distance of the gradients from its expected value per unit standard deviation measures how rare the current occurrence is [eq-\ref{eq1}]. We multiply this distance as a scalar weight with each of the gradients [eq-\ref{eq2}]. For sparse values this distance will be large, the opposite is also true. It practically acts like a low pass filter thus passes the desired signals with high amplitude and dampens the rest. For maintaining a stable bound to this amplification factor, we cut it off from $1/\rho$ to $\rho$ range. Empirically $\rho=3$ found to be effective in our experiments. The method can be used with or without momentum. SGDM (SGD with momentum) always demonstrates superior performance than vanilla SGD \cite{liu2020improved}. SGDM comes with high resistance to sudden change. It makes the process robust to stochastic noise but diminishes sparse change as well. This phenomenon is a widely researched issue\cite{kingma2014adam, duchi2011adaptive}. Since our method is specially designed to enforce rare signal, it makes a effective combination with momentum. The potential of our method to rescue rare signal with momentum proves the effectiveness of the claim in later section.
For gradients - $g_1, g_2, g_3 ..., g_t $ with mean $\mu_t$, standard deviation $\sigma_t$ and the constant $\ \rho>1 $ we can define the distance operator $\Delta_{\rho}$ in eq-[\ref{eq1}].
\begin{equation}
\begin{split}
\ \Delta_{\rho} ( g_{t} ,\mu _{t} ,\sigma _{t} \ ) \ =\Bigl\{\ \begin{array}{ l c }
min( abs( g_{t} -\mu _{t}) /\sigma _{t} ,\rho )g_t & if,\ abs( g_{t} -\mu _{t}) /\sigma _{t} \  >\ 1\\
max( abs( g_{t} -\mu _{t}) /\sigma _{t} ,1/\rho )g_t & otherwise
\end{array}\\
\label{eq1}
\end{split}
\end{equation}
\begin{equation}
\begin{split}
\mu_t    & =  \mathbb{E}(g_{i}) ; \sigma_t  =  \sqrt{\mathbb{E}(g_{i}^2) - \mathbb{E}(g_{i})^2} ; i = {t-1 , t-2, ... t-n}\\ 
m_t      & =  beta * m_{t-1} + \Delta_{\rho}(g_t,\mu_t,\sigma_t) \\
\theta_t & =  \theta_{t-1} - \alpha * m_t \\
\label{eq2}
\end{split}
\end{equation}
We define the following function as a standard template of a sparse signal generator for our analysis. Based on various values of $C$ , $u$ and $N$ we will examine the behaviour of different optimization techniques where $N>=3$.
\begin{equation}
\mathrm{f}(t) = \left\{ \begin{array}{ll} C, & \mathrm{if} \ {t \mathbin{\%} N = {0}} \\ u, & \mathrm{otherwise} \end{array} \right.
\label{eq3}    
\end{equation}
\begin{lemma}
\label{lemma1}
 Using gradients generated from eq. [\ref{eq3}] the consecutive values for t=1 to N would be - $g_1=g_2 ... =g_{N-1} = u ; \ g_N = C $. Initializing momentum update equation $m_{t+1} = \beta m_t + m_{t+1}$, with $m_0 = 0$, $kN^{th}$ momentum will be -
\begin{equation*}
m_{kN} \ =\ \ \beta _{k}^{N}( u\beta \beta _{N-1} \ +C) \ \ \ \ \ \ \ \ \ where,\ \beta _{x} \ =\frac{\beta ^{x} -1}{\beta -1} \ \ \ \ \ ;\ \ \ k\  >\ 0\ and\ k\ \epsilon \ \mathbb{Z} \ \ \ \ \ \ \ \ \ \ \ \ \ \ \ \ \ \ \ \ \ \ \ \ \ \ \ \ \ \ \ \ \ \ \ \ \ \ \ \ \ \ \ \ \ \ \ \ \ \ \ \ \ \ \ \ \ 
\end{equation*}
\end{lemma}

\begin{figure}[h!]
\vskip 0in
\centering
\includegraphics[width=10cm]{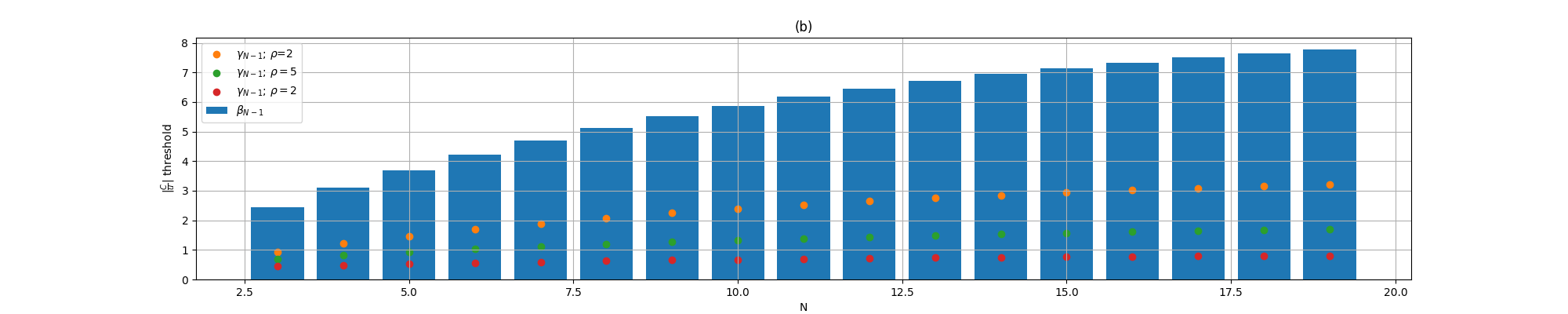}
\caption{$\beta_N \ , \gamma_N \ for \ \rho=2,3,5$ is shown for N steps. }
\label{fig2}
\vskip -0in
\end{figure}

If $uC<0$ (u and C with opposing sign) the effect of momentum will be most destructive. In subsequent sections we will consider this worst case scenerio for our analysis. From [\ref{lemma1}] $m_N$ to follow the direction of C the criterion is as follows -
\begin{equation}
\begin{split}
 |u \beta \beta_{N-1}| < |C| \ \ \ 
 => \ |\frac{C}{u}| > \beta \beta_{N-1}
\end{split}
\label{eq4}
\end{equation}
Fo example, a typical value of $\beta=0.9$ and N=3 the term $|\frac{C}{u}|>|\beta * \beta_{N-1}| = 2.44$, for N=9 it grows up to 5.51 Figure [\ref{fig2}b].
For N = 9 , $\beta * \beta_{N-1} > 5.5$. Sparse signals below this threshold follows the direction of u adversely, shown in Figure [\ref{fig2}a]. In contrast, here GQ will boost the sparse gradient, pushing this limit at least $\rho$ times lower, up to $\approx\rho^2$ depending on how small $qlen$ is compared to sparse frequency N.
\begin{lemma}
\label{lemma2}
For a queue of length L composed of two unique elements u and C where number of u is L-1. Then $\Delta_{\rho}(u) = \phi u$ ; where $\phi \ =max\left(\frac{1}{\ \sqrt{( L-1)} \ \ \ } ,\frac{1}{\rho }\right)$.
\end{lemma}
\begin{lemma}
\label{lemma3}
 Using the same condition of [\ref{lemma1}] momentum at $kN^{th}$ step boosted  with [\ref{eq1}],
\begin{gather*}
\Delta _{\rho }( m_{kN}) = \ \beta^{N(k-1)}\left( u \beta \gamma _{N-1}^{0}\ +\ \rho C\right) \ + \ \beta_{k-1}^N \left( u \beta \gamma _{N-1} \ +\ \rho C\right) \ \\
where,\ \gamma _{N-1}^{0} \ =\beta ^{N-1-L} \beta _{L} +\ \beta _{( N-1-L)}\frac{1}{\rho } \ \ ;\ \ \ \gamma _{N-1} \ =\phi \beta ^{N-1-L} \beta _{L} +\ \beta _{( N-1-L)}\frac{1}{\rho } \ 
\end{gather*}
\end{lemma}
 $\gamma_x^0 > \gamma_x$ so if the first term follows direction of C the 2nd term will do as well. So, for gq boosted momentum the $|C/u|$ bound for $\Delta_{\rho}(m_{kN})$ to follow the direction of "C"  will be -
\begin{equation}
\begin{split}
|u \beta \gamma_{N-1}^0| < |\rho C| \ \ \ \
=> \ |\frac{C}{u}| > \frac{1}{\rho} (\beta \gamma_{N-1}^0)
\label{eq6}
\end{split}
\end{equation}
 Since $\rho>1$, $\gamma_{x}^0 = (\beta^{x-L} \beta_{L} + \beta_{x-L} \frac{1}{\rho}) < (\beta^{x-L} \beta_{L} + \beta_{x-L}) = \beta_{x} $.
 if $\ \frac{L}{N} << 1 \, \ \gamma_x^0 \approx \frac{\beta_x}{\rho}$\\
  Comparing the $|C/u|$ lower bounds of eq[\ref{eq4}] and eq[\ref{eq6}] - 
\begin{equation}
\begin{split}
\frac{1}{\rho} \beta \gamma_{N-1}^0 < \beta \beta_{N-1} \\
if \ \frac{L}{N} << 1 \ ; \ \gamma_{N-1}^0 \rightarrow \frac{\beta_{N-1}}{\rho} \ ; \ 
\frac{1}{\rho^2} \beta \gamma_{N-1}^0 << \beta \beta_{N-1}
\label{eq7}
\end{split}
\end{equation}

So, gq boosting lowers the $|C/u|$ limit for $m_N$ to follow the direction of 'C' at least $\rho$ times to  $~\rho^2$ times based on the $L/N$ ratio.

\begin{figure}[h!]
\vskip 0in
\centering
\includegraphics[width=10cm]{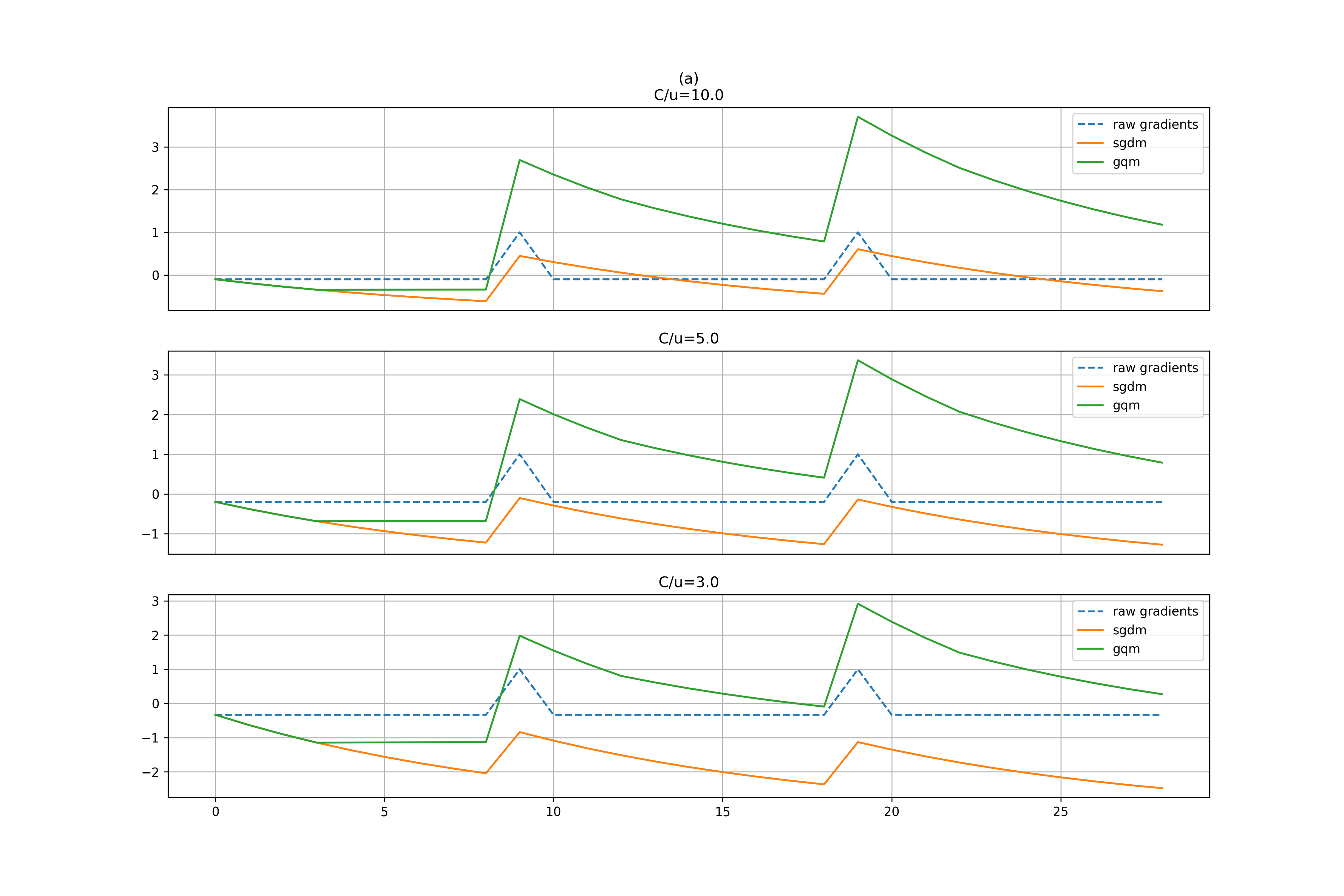}
\caption{Momentum values for gradient generated from [\ref{eq3}].}
\label{fig3}
\vskip -0in
\end{figure}

We have found too large queue length to be less usefull in our experiments. After several updates the weight will be very different. So gradients wrt older weights will not be a good indicator of the current the trend. We found 3 to 5 queue length good enough for our purpose. We also introduced a variable queue length scheme by monitoring the loss convergence pattern bounded by a upper bound. The more steps back loss pattern shows continuous convergence the higher the queue length will be. A small window is slided over the queue of loss from current step t up to the t-qlen step. It continues to move back as far as the sum within the window increases. It is a measure of how long the loss has been decreasing. The window gives it robustness to noise.


\subsection{Monotonous and Noisy Gradients}
\begin{figure}[h!]
\vskip 0in
\centering
\includegraphics[width=18cm]{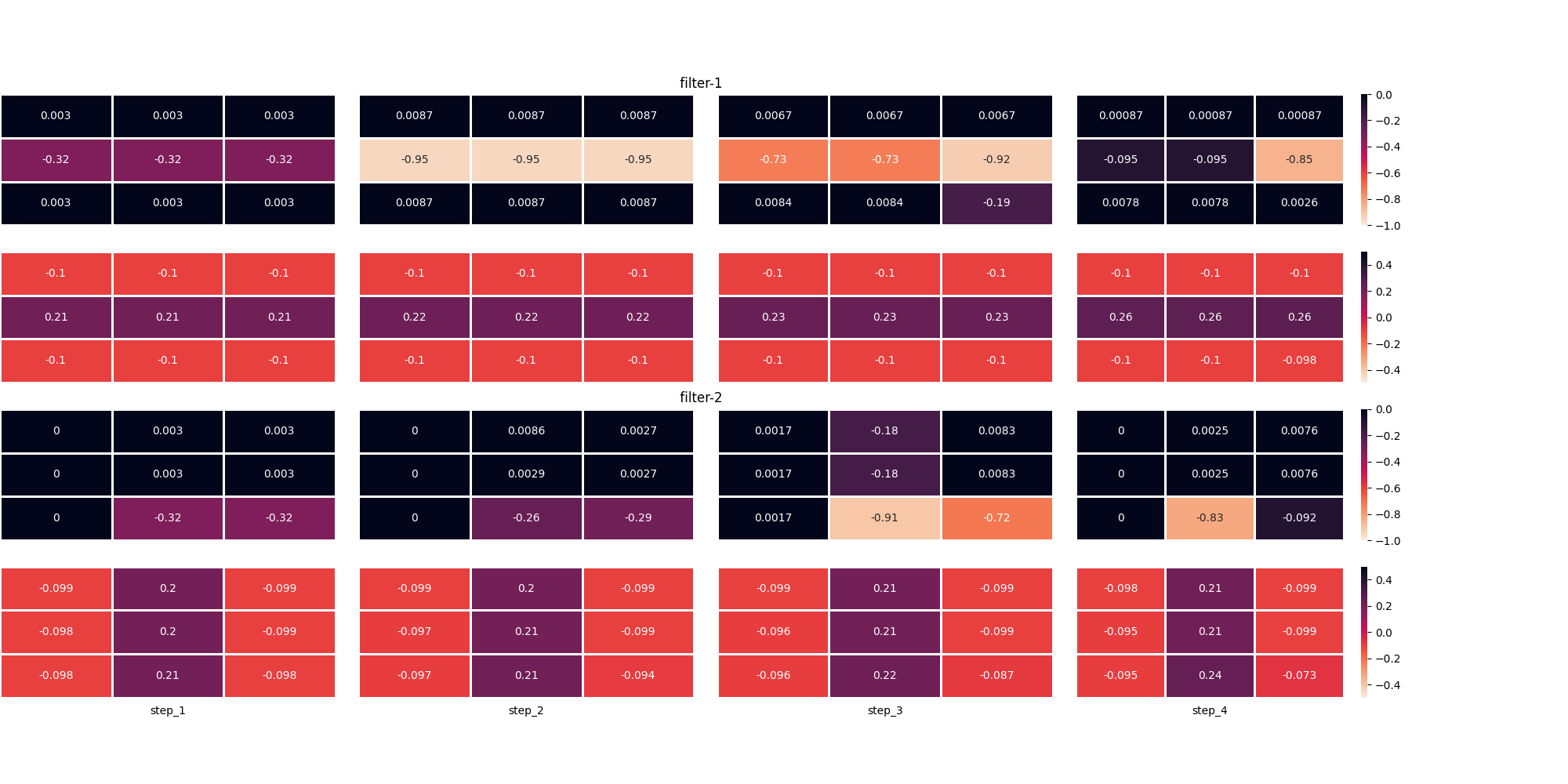}
\caption{Gradients and propagation of weights for 5 consequitive time steps for filter-1 and filter-2 trained on synthetic line dataset}
\label{fig4}
\vskip -0in
\end{figure}

We illustrate a problem of line detection for our subsequent analysis. Assume a random uniform pull of B samples resulting in “p” horizontal and “q” with vertical line  samples where, $p>>q$ and $p+q = B$. The detector model has 2 parallel $3x3$ CNN (convolutional Neural Network) filters at its input layer followed by respective global max pooling layers and a common dense layer with 2-input and 1-output node at the end for class prediction. Based on their initialization one $3x3$ will start learning the horizontal lines detection and another will learn vertical ones. We name them filter-1 and filter-2 respectively. Each filter will have the largest gradients for samples containing the feature it is learning. So horizontal lines will result in the largest gradients for filter-1 while filter-2 will also have some non-zero gradients driving it far from its optimal parameters. This is the component that can act as a destructive monotonous counterpart for these filter-2 parameters. Below the optimal values for filter-1 and 2. \begin{gather*}
filter-1 \ =\begin{matrix}
-1 & -1 & -1\\
\ \ 2 & \ \ 2 & \ \ 2\\
-1 & -1 & -1
\end{matrix} \ \ \ 
filter-2 \ =\begin{matrix}
-1 & 2 & -1\\
-1 & 2 & -1\\
-1 & 2 & -1
\end{matrix}
\end{gather*}
Due to the abundance of samples from "p", it will frequently cause beneficial but redundant updates for filter-1 while corrupting  filter-2 weights. Rarely occurring “q” samples will ignite the filter-2 kernel which are the sparse updates of interest. Evidently, suppressing the p updates for making the q (sparse) gradients effective is the solution for filter-2 to learn properly, which is what we are going to do later in this work.
Apart from above, there can be generic noise in gradients during stochastic updates. Presence of any irrelevant feature in the sample (e.g noisy background) will add some noisy component in respective gradients. Choosing a reasonably large mini-batch can be an easy escape from this situation. Noises come uniformly with every possible pattern thus canceling each other's effect when averaged. Many of the existing works tried to deal with the stochastic coming from per sample noise [][][] others focused on enforcing the sparse gradients by updating the optimization method [][][] or by choosing diverse training samples [][][].
We implemented this problem with a synthetic dataset of lines. We simulated with 95 horizontal line and 5 vertical line and the same model described above. Figure-[\ref{fig4}] shows the gradients and propagation of weights for 4 consecutive time steps. Note here filter-1 expects negative gradient at 2nd row while filter-2 expects the same for 2nd column. For other row or columns the opposite is expected. Only the mini-batch of step_3 contained a sparse update (vertical line). It is clear from the figure that filter-1 is learning the horizontal line detection quite fast. Filter-2 useful gradients at  step_3. At other steps filter-2 is having non opposing gradients which is acting as destructive noise from the monotonous updates. For the last two values of the 3rd row of the of filter-2 gradients the destructive interference is quite high. 

\begin{figure}[h!]
\vskip 0in
\centering
\includegraphics[width=18cm]{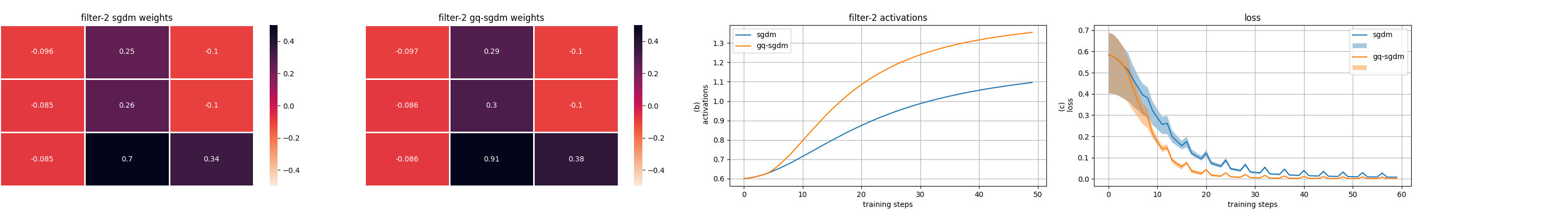}
\caption{For sgdm and gq-sgdm method (a) weights of filter-2 (b) activations of filter-2 with training steps (c) loss }
\label{fig5}
\vskip -0in
\end{figure}

Online SGD or small mini-batch suffers from the noises from individual samples. If we keep increasing the batch size the monotonous gradients will become dominant enough to suppress the sparse ones. \\
\begin{asmp}
\label{asmp1} 
From the above hypothesis we can conclude, optimal batch size "B" stays within the following bounds -\\
i. Lower bound [$\omega$] : It should be large enough to overcome the stochastic noises.\\
ii. Upper bound [$\psi$] : It should be small enough for preserving the sparse updates.\\
\end{asmp}
For gradients pulled from “p” and “q” samples -
$\displaystyle \mathbb{E}\left( g^{q}\right) \ =\frac{1}{q}\sum _{i\ \epsilon \ q} g_{i}^{q} \  and \ \mathbb{E}\left( g^{p}\right) \ =\ \frac{1}{p} \ \sum _{i\ \epsilon \ p} g_{i}^{p}$.\\ In destructive cases where $\mathbb{E}(g^q) , \ \mathbb{E}(g^p)$ have opposing signs and $||\mathbb{E}(g^q) / \mathbb{E}(g^p)|| <= p/q$, any uniform mini-batch gradient will diminish the sparse update completely hence will cross the upper bound $\psi$. The solution is to boost $\mathbb{E}(g^q) / \mathbb{E}(g^p)$, so that we can select a mini-batch large enough to cross the lower bound $\omega$. \\
\begin{asmp}
\label{asmp2}
eq[\ref{eq1}] imposes a minimal distance limit of 1 standard deviation for a signal to be sparse. Ideally, $|\mathbb{E}(g^q)|>>|\mu+\sigma|$ and $\mathbb{E}(g^p)\approx\mu$. So there exists a $\rho$ for which,
$|\mathbb{E}(g_{q}) |\  >\ |\mu _{g} \ +\ \rho \sigma | \ , \ |\mathbb{E}(g_{p}) |\ < \ |\mu _{g} \ +\ \sigma /\rho |$ where $\rho>1 \ and \ \rho\epsilon\mathbb{Z}$
\end{asmp}
From [eq-\ref{eq1}] ,
\begin{gather*}
\Delta_{\rho} \left( \mathbb{E}(g^{q})\right) \ =\ \rho \mathbb{E}(g^{q}) \ ; \ \ \
\Delta_{\rho} \left( \mathbb{E}(g^{p})\right) \ =\ \mathbb{E}(g^{p}) /\rho 
\end{gather*}
	[\cite{friedlander2012hybrid}] derived the following convergence equation for gradient descent with error in gradient calculation -
\begin{equation}
f( x_{k} +1) \ −\ f\left( x^{*}\right) \ \leq \ ( 1\ −\ \mu /L)\left[ f( x_{k}) \ −\ f\left( x^{*}\right)\right]\\
\ +\frac{1}{2L} \ \| e_{k} \| ^{2}
\label{eq8}
\end{equation}
assuming, $f(x)$ is strongly convex with positive $\mu$ and L-Lipschitz continuous. Hypothetically, a batch containing "q" type samples would be fully effective if it can preserve the sparse magnitude i.e $\mathbb{E}(g^b)=\mathbb{E}(g^q)$ without any interference. So, the error $||e_k||$ in [\ref{eq8}] can be modeled as deviation of $\mathbb{E}(g^b)$ from $\mathbb{E}(g^q)$. Note, there can be another component of this error coming from stochistic noise. We are aiming to improve accuracy on optimal and beyond optimal batch size above the lower bound $\omega$ of [\ref{asmp1}]. So this component can be safely overlooked for this study.
\begin{gather*}
\begin{aligned}
||e_{k} ||=\ ||\mathbb{E}\left( g^{q}\right) \ -\ \ \mathbb{E}\left( g^{b}\right) ||\ \ ,\ where\ ||\mathbb{E}\left( g^{q}\right) ||\  >\ ||\mathbb{E}\left( g^{b}\right) ||\  >\ ||\mathbb{E}\left( g^{p}\right) ||\ \ \\
\mathbb{E}\left( g^{b}\right) \ =\ 1/B\left(\sum _{i\ \epsilon \ q} g_{i}^{q} \ +\ \sum _{i\ \epsilon \ p} g_{i}^{p}\right) =\ 1/B\ \left( q\mathbb{E}\left( g^{q}\right) \ +\ p\mathbb{E}\left( g^{p}\right)\right) \ \ \ \ \ 
\end{aligned}\\
\begin{aligned}
 &  & \\
case\ 1:\ \ ||\mathbb{E}\left( g^{q}\right) /\mathbb{E}\left( g^{p}\right) ||\ \gg \ \ p/q; & \ \mathbb{E}\left( g^{b}\right)\rightarrow \mathbb{E}\left( g^{q}\right) \ ; & ||e_{k} ||\ =\ 0\ \ \ \ \ \ \ \ \ \ \ \ \ \ \ \ \ \ \ \ \ \ \ \ \ \ \ \ \ \ \ \ \ \ \ \ \ \\
case\ 2:\ \ ||\mathbb{E}\left( g^{q}\right) /\mathbb{E}\left( g^{p}\right) ||\ =\ \ p/q;\  & \ \mathbb{E}\left( g^{b}\right) =0\ \ \ \ \ \ \ \ ; & ||e_{k} ||\ =\ ||\mathbb{E}\left( g^{q}\right) ||\ \ \ \ \ \ \ \ \ \ \ \ \ \ \ \ \ \ \ \ \ \ \ \ \\
 &  & where,\ \ \mathbb{E}\left( g^{q}\right) \ *\ \mathbb{E}\left( g^{p}\right) \ < 0\ \ \ \ \ \ \\
case\ 3:\ ||\mathbb{E}\left( g^{q}\right) /\mathbb{E}\left( g^{p}\right) ||\ \ll \ p/q;\ \  & \ \mathbb{E}\left( g^{b}\right)\rightarrow \mathbb{E}\left( g^{p}\right) ;\  & ||e_{k} ||=\ ||\mathbb{E}\left( g^{q}\right) \ -\ \ \mathbb{E}\left( g^{p}\right) ||\ \ \ \ \ \ \ 
\end{aligned}\\
\ 
\end{gather*} 
Note, in case-2 we demonstrated the worst outcome only. Case-2 and 3 exceeds the upper bound $\psi$ of assumption[\ref{asmp1}]  If assumption[eq-\ref{asmp2}] holds, applying $\Delta_{\rho}$ to $\mathbb{E}\left( g^{q}\right) /\mathbb{E}\left( g^{p}\right)$ both the cases can be turned into case-1 for a large enough $\rho$.

\begin{equation*}
\begin{aligned}
\Delta _{\rho }\left( ||\mathbb{E}\left( g^{q}\right) /\mathbb{E}\left( g^{p}\right) ||\right) =\rho ^{2} \ ||\mathbb{E}\left( g^{q}\right) /\mathbb{E}\left( g^{p}\right) ||\\
if\ \rho \ is\ high\ enough\ ,\ \ \ \ \ \ \ \ \ \ \ \ \ \ \ \ \ \ \ \ \ \ \ \ \ \ \ \ \ \ \ \ \ \ \ \ \ \ \ \ \ \ \ \\
\rho ^{2} \ ||\mathbb{E}\left( g^{q}\right) /\mathbb{E}\left( g^{p}\right) ||\ \gg \ p/q\ \ \ \ \ \ \ \ \ \ \ \ \ \ \\
\Delta _{\rho }\left(\mathbb{E}\left( g^{b}\right)\right)\rightarrow \mathbb{E}\left( g^{q}\right) ;\ ||e_{k} ||\rightarrow 0\ \ \ \ \ \ \ \ \ \ \ \ \ \ \ \ \ \ \ \ \ \ \ \\
Here\ ,\ 1\ < \rho \ \leqslant \ \zeta \ \ \ \ \ \ \ \ \ \ \ \ \ \ \ \ \ \ \ \ \ \ \ \ \ \ \ \ \ \ \ \ \ \ \ \ \ \ \ \ \ \ \ \ \ \ \\
 \\
\end{aligned}\begin{aligned}
For\ \ \rho \ =\ \zeta \ ,\ \Delta _{\zeta }\mathbb{E}\left( g^{b}\right) \ =\ \mathbb{E}\left( g^{q}\right) \ \ \ \ \ \ \ \ \\
\Delta \zeta \left( 1/B\ \left( q\mathbb{E}\left( g^{q}\right) \ +\ p\mathbb{E}\left( g^{p}\right)\right)\right) =\mathbb{E}\left( g^{q}\right) \ \ \ \ \ \ \ \ \ \\
\ \ \ q\zeta \mathbb{E}\left( g^{q}\right) \ +\ p\mathbb{E}\left( g^{p}\right) /\zeta =B\mathbb{E}\left( g^{q}\right) \ \ \ \ \ \ \\
\zeta ^{2} q\mathbb{E}\left( g^{q}\right) \ -\zeta B\mathbb{E}\left( g^{q}\right) \ +\ p\mathbb{E}\left( g^{p}\right) =0\ \ \\
\zeta \ =\frac{B\mathbb{E}\left( g^{q}\right) \ +\ \sqrt{B^{2}\left(\mathbb{E}\left( g^{q}\right)\right)^{2} -\ 4q\mathbb{E}\left( g^{q}\right) p\mathbb{E}\left( g^{p}\right)}}{2q\mathbb{E}\left( g^{q}\right)}
\end{aligned} \ \ \ \ \ \ \ \ \ \ \ \ \ \ \ \ \ \ \ \ \ \ \ \ \ \ \ \ \ \ \ \ \ \ \ \ \ \ \ 
\end{equation*}
\subsection{Enhancing Intra-Batch sparsity}
In the above analysis we see that a data sample can be specifically responsible for teaching the model some particular sub-task/tasks (e.g horizontal or vertical line detection) to achieve the collective objective (e.g line detection).  Each sub-task can be learnt by a subset of the parameter space (like the two convolutional filters in our case). If the sample contains a rare feature, blending its gradient with regular ones can make the model to not learn the associated sub-goal. The problem here is, as the group gets larger the internal sub-tasks will start to differ and interfere with each other destructively as we have seen in Figure [\ref{fig4}]. It can create enough interference with sparse updates to get it nullified by the monotonous counterparts, pursuing different sub-goals. It is the reason for failure of too high mini-batch size to reach good accuracy []. To handle this we further cluster data samples within a large batch. It is done based on the similarity in the inherent feature space of each sample. So the sub objectives within each cluster stays aligned. Consequently, samples with sparse objectives are expected to stay in similar clusters.

For a particular parameter $W$ of a model. The mini-batch size is B and the number of classes is C. At every update we will get a $BxC$ matrices of loss hence a matrix of $BxC$ gradients.
Online SGD will take the overall mean of $m*n$ gradients to update $W$. We do the following instead -\\
- Cluster the n samples into $k$  groups, replace each cluster with their center weighted by group population, resulting in $KxC$ gradients.\\
- Take average across the class dimension, resulting in $K$ gradients for weight $W$.\\
- Calculate the K distances  $\Delta$ element wise for each of the $K$ gradients.\\
- Take the weighted mean as follows - $ G*\ =\ \sum _{i=1}^{k} \ \upDelta _{i} \ *\ G_{i}$

 Image search or re-identification algorithms[][] uses feature map F from an intermediate dense layer as the matching criteria. Inspiring from that we extract the vector F from an intermediate dense layer of the model. For a batch total B the samples are clustered into k clusters using the corresponding feature vectors. Now within each of these clusters we squeeze out the unnecessary information by applying our “sparsity amplification operator” on the cluster center. 
\begin{algorithm}
\caption{Algorithm (GQ)}
\begin{algorithmic} 
\STATE $Init$ $\theta_0,m_0$
\STATE $Q= [0,0,...],L = length(Q)$
\STATE $set \ \alpha, \beta, \rho, k $
\WHILE{$epoch$}
\STATE $\mu^{t} \ =\frac{1}{L} \ \sum _{i=t-1}^{t-L} g_{i} \ ;\ \ \sigma^{t} \ =\ \frac{1}{L} \ \sum _{i=t-1}^{t-L}( g_{i} \ -\ \mu _{t})$
\FOR{$x_b^t,y_b^t$  in \textbf{Batches}}
\STATE $z_b^t \leftarrow \mathbb{F}_{fp}(x_b^t,y_b^t,\theta^t), \ \ $; $z_b^t$= feature vector $\mathbb{F}_{fp}$=Forward pass.
\STATE $g_b^t \leftarrow \mathbb{F}_{bp}(x_b^t,y_b^t,\theta^t), \ \ $; $g_b^t$ = batch gradients, $\mathbb{F}_{bp}$ = backward pass.
\STATE $g_1, g_2 ...g_k = \mathbb{C}(z_b^t,g_b^t,k) \ $; $\mathbb{C}$ = Apply KMeans on $z_b^t$ with K=k, split $g_b^t$ based on it.
\STATE $g^* = \frac{1}{B} \sum_{i=1}^{k} len(g_k) * \Delta_{\rho}(mean(g_k),\mu^t,\sigma^t)$
\STATE $\theta_{t+1} \leftarrow \theta_t+\alpha g^*,$
\STATE $Q    \leftarrow g_{t}$
\ENDFOR
\ENDWHILE
\end{algorithmic}
\end{algorithm}
\vspace*{-0.05in}
\section{Experimental Results}
We have experimented on CIFAR10, MNIST dataset with SGDM(Stochistic Gradient Descent with Momentum) , sgdm boosted with our GQ(grad queue) method namely GQ-SGDM. For GQ-SGDM we have tried with one single cluster and multiple clusters for different batch sizes. For reuters dataset we have used ADAM optimizer and shown comparison with ADAM boosted with GQ namely GQ-ADAM. In every cases grad queue boosted method out performed vanilla optimizers and higher number of cluster out performs single cluster for large batches. For beyound optimal batch sizes we have used number of clusters equals to its ratio with optimal size. 
\begin{figure}[h!]
\vskip 0in
\centering
\includegraphics[width=14cm]{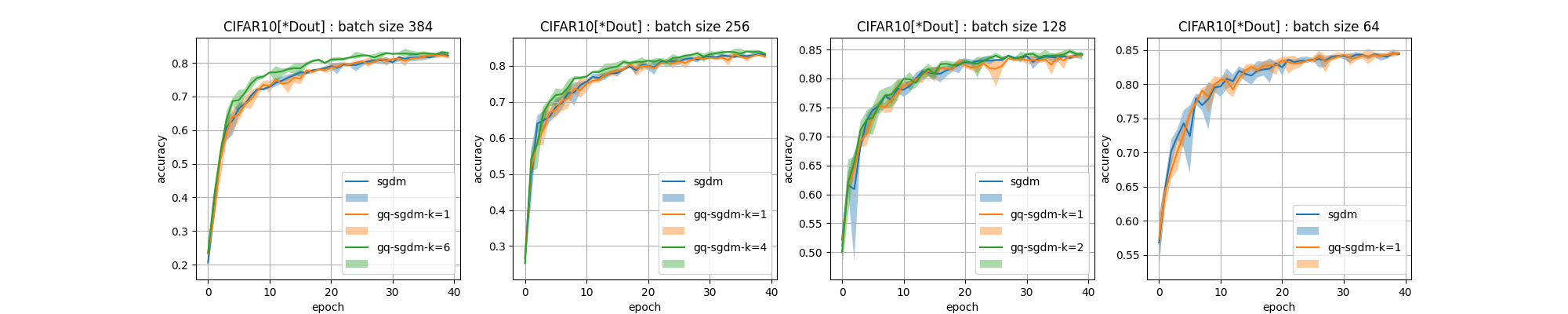}
\includegraphics[width=14cm]{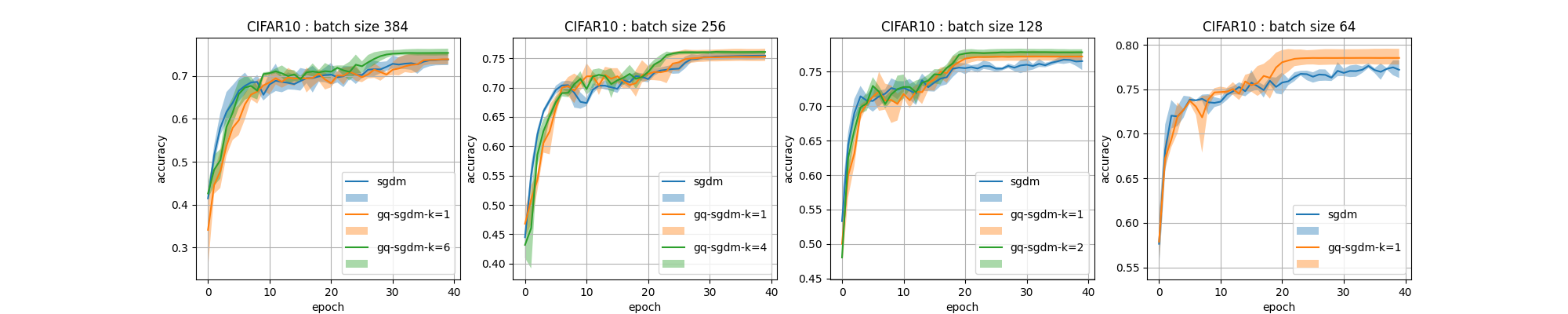}
\includegraphics[width=14cm]{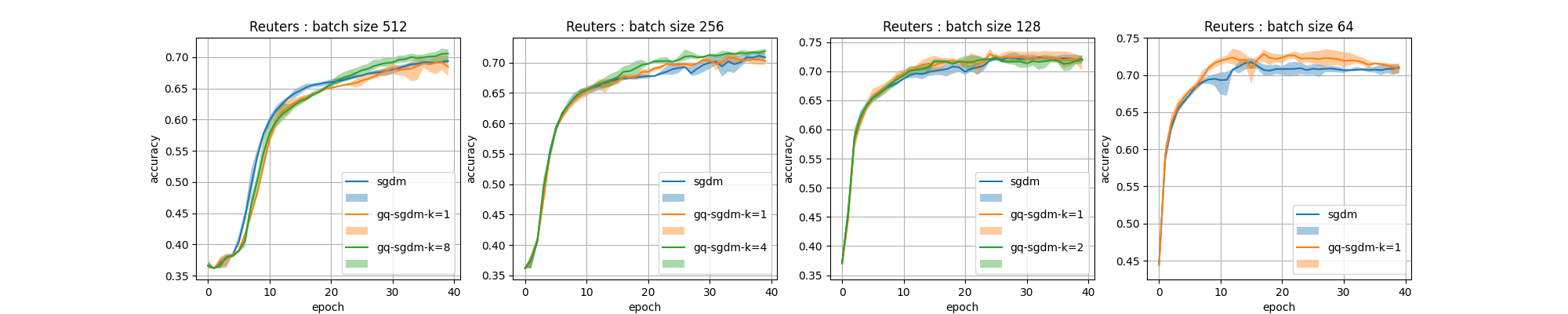}
\includegraphics[width=14cm]{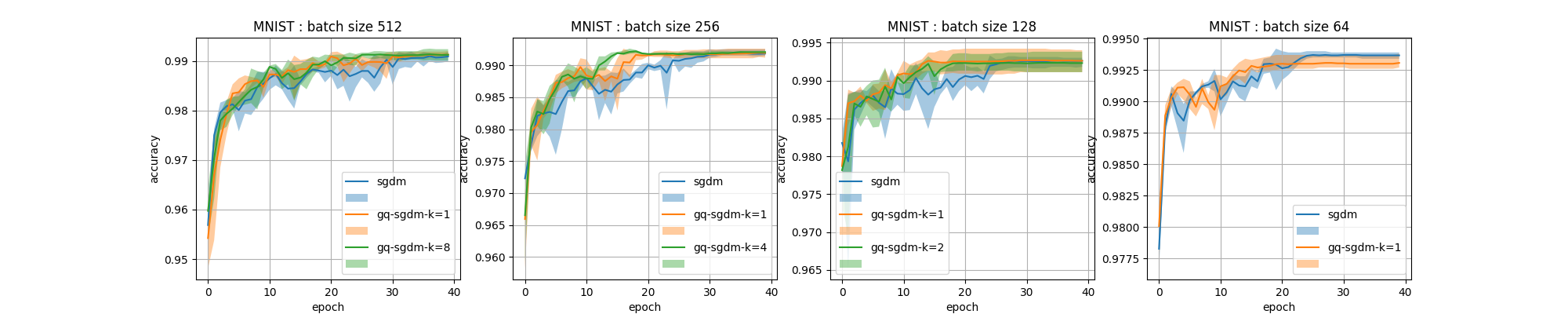}
\caption{Results on CIFAR10 \cite{krizhevsky2010cifar}, Reuters \cite{lewis1997reuters}, MNIST \cite{deng2012mnist} dataset}
\label{fig1}
\vskip -0in
\end{figure}
\label{results}

\clearpage
\section{Conclusion}
We demonstrate the potential of emphasizing on gradients having more potential on training time. We achieve it by maintaining a active queue of gradients with time. Using our approach classification task was performed in several widely used datasets in computer vision and natural language processing domain. Robust techniques to extract the impactful gradients can be explored further while memory and run-time can be optimized for making the method widely applicable.
\appendix
\section*{Appendix A.}
\textbf{Proof of Lemma \ref{lemma1}}\\
Using the momentum equation with $m_0=0$, $k >0$ and $k \epsilon \mathbb{Z}$.
\begin{align*}
m_{x} \ \ \ \ \  & =\beta \ m_{x-1} \ _{\ } +\ g_{x} \ =\ \beta ^{2} m_{x-2} \ +\ \beta g_{x-1} \ +\ g_{x} \ \\
m_{x} \ \ \ \ \  & =\ \beta ^{x} m_{0} \ +\ \beta ^{x-1} g_{1} \ +\beta ^{x-2} g_{2} \ +\ .\ .\ .\ +\ \beta g_{x-1} \ +\ g_{x} \ \\
m_{N-1} \  & =\beta ^{N-1} m_{0} \ +\ \beta ^{N-2} \ g_{1} \ +\ \beta ^{N-3} \ g_{2} \ +\ ...\ +\ \beta g_{N-2} \ +\ g_{N-1}\\
 & =\ \beta ^{N-1} m_{0} \ +\ u\ \left[ \beta ^{N-2} \ \ +\ \beta ^{N-3} \ \ +\ ...\ +\ \beta \ +\ 1\right] \ =\beta ^{N-1} m_{0} \ +\ u\ \ \frac{\beta ^{N-1} -1}{\beta -1}
 \end{align*} 
\begin{equation}
\begin{aligned}
 m_{N-1} &=\ \beta ^{N-1} m_{0} \ +\ u\ \beta _{N-1} \ \ ,\ \ taking\ \ \beta _{x} \ =\frac{\beta ^{x} -1}{\beta -1} \\
 m_{x} &=\ \beta ^{x} m_{0} \ +\ u\ \beta _{x} \ \ ,\ \ taking\ \ \beta _{x} \ =\frac{\beta ^{x} -1}{\beta -1} \ \ \ \ \ \ \ \ \ \ \ \ \ \
 \ \ \ \ \ \ \ \ \ \ \ \ \ \ \ \ \ \ \ \ \ \ \ \ \ \ \ \ \ \ \ \ \ \
\end{aligned}
\label{eq9}
\end{equation}
Eq. [\ref{eq9}] can be used as a general formula for expanding the momentum values with time step, starting from initial value $m_0$ and expanding for the next x steps while u is repeated every time.
\begin{align*}
m_{N} \ \ \ \  & =\beta m_{N-1} \ +\ C\ =\ \beta ^{N} m_{0} \ +\ u\beta \beta _{N-1} \ +\ C\ \ \ \ \ \ \ \ \ \ \ \ \ \ \ \ \ \ \ \ \ \ \ \ \ \ \ \ \ \ \ \ \ \ \ \ \ \ \ \ \ \ \ \ \ \ \ \ \ \ \ \ \ \ \ \ \ \ \ \ \ \ \ \ \ \ \ \ \ \ \ \ \ \ \ \ \\
 & \ =\ u\beta \beta _{N-1} \ +\ C\ ,\ with\ m_{0} =0\\
m_{2N-1} & =\ \beta ^{N-1} m_{N} \ +\ u\beta _{N-1}\\
m_{2N} \ \ \  & \ =\ \beta m_{2N-1} \ +\ C\ =\ \beta ^{N} m_{N} \ +\ u\beta \beta _{N-1} \ +\ C\\
 & \ \ =\ \left( \beta ^{N} +1\right)( u\beta \beta _{N-1} \ \ +\ C)\\
m_{3N} \ \ \  & \ =\beta \left( \beta ^{N-1}\left(\left( \beta ^{N} +1\right)( u\beta \beta _{N-1} \ \ +\ C)\right) +\ \beta _{N-1} u\right) \ +\ C\ \\
 & \ =\ \left( \beta ^{2N} +\beta ^{N} +1\right)( \beta \beta _{N-1} u+C) \ \\
 & \ =\beta _{3}^{N} \ ( u\beta \beta _{N-1} +\ C) \ \\
m_{kN} \ \ \  & \ =\beta _{k}^{N} \ ( u\beta \beta _{N-1} +\ C) \ 
\end{align*}

\textbf{Proof of Lemma 3.2}
\begin{gather*}
\ \begin{aligned}
\sigma  & =\ \sqrt{\mathbb{E}\left( x^{2}\right) \ -\ \mathbb{E}( x)} \ \ \ =\ \sqrt{\frac{(L-1) u^{2} +C^{2}}{L} \ -\ \left(\frac{( L-1) u\ +\ C}{L}\right)^{2}} \ \ \ =\ \ \sqrt{( L-1)} \ \left(\frac{u-C}{L}\right) \ \ \ \ \ \ \ \ \ \ \ \ \ \ \ \ \ \ \ \ \ \ \ \ \ \ \\
\Delta ( u) & \ =\ \frac{u-\mu }{\sigma } u=\frac{\ u\ -\ \frac{( L-1) u\ +\ C}{L}}{\sigma } u\ =\frac{u}{\ \sqrt{( L-1)} \ \ \ }\\
\Delta _{\rho }( u) & =max\left(\frac{1}{\ \sqrt{( L-1)} \ \ \ } ,\frac{1}{\rho }\right) \ u\ \ \ \ \ \ \ \ \ \ \ \ \ \ \ \ \ \ \\
 & taking,\phi \ =max\left(\frac{1}{\ \sqrt{( L-1)} \ \ \ } ,\frac{1}{\rho }\right) \ \ \\
since, & \rho  >1\ and\ L >3\ ;\ \phi < 1
\end{aligned}
\end{gather*}\\
\textbf{Proof of Lemma 3.3}\\
Let's consider a zero initiated queue of length L is being filled with values from [\ref{eq3}]. Boosting with eq. [\ref{eq1}] is possible if the queue is filled after initial L steps. \\
For $0 < t < L$, from eq. [{\ref{eq9}] with $ m_0=0 \ , \ m_L=\beta u$ and $\Delta_{\rho}(u)=u$\\
For $t>L$, using Lemma \ref{lemma2}, $\Delta_{\rho}(u)=\phi u$


From eq. [\ref{eq9}] -
\begin{gather*}
\begin{aligned}
Starting\ \ \  & from\ t=0\ \ and\ expanding\ up\ to\ x^{th} \ step,\\
m_{x} \ \ \ \ \ \ \ \ \ \ \ \  & =\ \beta ^{x} m_{0} \ +\ \beta _{x} u\ \ [ here,\ x >L]\\
Starting\ \ \  & from\ L^{th} \ ( \ where,\ x >L) \ step\ \ and\ expanding\ for\ remaining\ x-L\ steps,\\
m_{x} \ \ \ \ \ \ \ \ \ \ \ \  & =\ \beta ^{x-L} m_{L} \ +\ u\beta _{x-L} \ \\
m_{N-1} \ \ \ \ \ \ \ \  & =\beta ^{( N-1-L)} m_{L} \ +\ u\beta _{N-1-L} \ =\ \beta ^{( N-1-L)} u\beta _{L} \ +\ u\beta _{N-1-L} \ \ [ here,\ N-1\  >\ L]\\
\Delta _{\rho }( m_{N-1}) \  & =\ \Delta _{\rho }\left( \ u\beta ^{( N-1-L)} \beta _{L} \ +\ u\beta _{N-1-L} \ \right) \ =\beta ^{N-1-L} u\beta _{L} \ +\ \beta _{( N-1-L)}\frac{u}{\rho }\\
 &  \begin{array}{l}
taking,\ \ \gamma _{x}^{0} \ =\beta ^{N-1-L} \beta _{L} +\ \beta _{( N-1-L)}\frac{1}{\rho } \ \ ,\ for\ 0< t\ < \ L\\
\ \ \ \ \ and\ \ \ \gamma _{x} \ =\phi \beta ^{N-1-L} \beta _{L} +\ \beta _{( N-1-L)}\frac{1}{\rho } \ \ ,\ for\ L< t\ ;\ \ 
\end{array}\\
\Delta _{\rho }( m_{N-1}) & =\ \gamma _{N-1}^{0} \ u\ \\
\Delta _{\rho }( m_{N}) \ \ \ \  & =\ u\beta \gamma _{N-1}^{0} \ +\ \rho C\\
\Delta _{\rho }( m_{2N-1}) & =\ \beta ^{N-1} m_{N} \ +\ u\gamma _{N-1} \ \\
\Delta _{\rho }( m_{2N}) \ \ \  & =\ \beta \left( \beta ^{N-1} m_{N} \ +\ u\gamma _{N-1} \ \right) \ +\ \rho C\ =\ \beta ^{N}\left( u\beta \gamma _{N-1}^{0} \ +\ \rho C\right) \ +\left( \ u\beta \gamma _{N-1} \ +\ \rho C\right)\\
\Delta _{\rho }( m_{3N-1}) & =\ \beta ^{N-1} m_{2N} \ +\ u\gamma _{N-1} \ \\
\Delta _{\rho }( m_{3N}) \ \ \  & =\ \beta \left( \beta ^{N-1} m_{2N} \ +\ u\gamma _{N-1} \ \right) \ +\ \rho C\ =\beta ^{2N}\left( u\beta \gamma _{N-1}^{0} \ +\ \rho C\right) \ +\ \left( \beta ^{N} +\ 1\right)\left( u\beta \gamma _{N-1} \ +\ \rho C\right)\\
\Delta _{\rho }( m_{kN}) \ \ \  & =\ \beta ^{( k-1) N}\left( u\beta \gamma _{N-1}^{0} \ +\ \rho C\right) \ +\ \left( \beta ^{( k-2) N} +\beta ^{( k-3) N} \ +...+1\right)\left( u\beta \gamma _{N-1} \ +\ \rho C\right)\\
 & =\ \ \beta ^{( k-1) N}\left( u\beta \gamma _{N-1}^{0} \ +\ \rho C\right) \ +\ \beta _{k-1}^{N}\left( u\beta \gamma _{N-1} \ +\ \rho C\right)
\end{aligned} \ \ \ \ \ \ \ \ \ \ \ \ \ \ \ \ \ \ \ \ \ \ \ \ \ \ \ \ \ 
\end{gather*}
\vskip 0.2in
\bibliography{sample}
\bibliographystyle{abbrvnat}

\end{document}